\documentclass[11pt, a4paper, logo, onecolumn,copyright]{googledeepmind}

\usepackage[authoryear, sort&compress, round]{natbib}
\title{Physical Autoregressive Model for Robotic Manipulation without Action Pretraining}

\renewcommand{\today}{}

\usepackage[utf8]{inputenc} 
\usepackage[T1]{fontenc}    
\usepackage{hyperref}       
\usepackage{url}            
\usepackage{booktabs}       
\usepackage{nicefrac}       
\usepackage{microtype}      
\usepackage{amsmath}
\usepackage{graphicx}
\usepackage{multicol}
\usepackage{siunitx}
\usepackage{array}
\usepackage[nameinlink]{cleveref}
\usepackage{bbm}
\usepackage{multirow}
\usepackage{subfig}
\usepackage{soul}
\usepackage{floatrow}
\usepackage{float}
\usepackage{wrapfig}
\usepackage{blindtext}
\usepackage{tablefootnote}
\usepackage{amsfonts}
\usepackage[flushleft]{threeparttable}
\usepackage{colortbl}
\usepackage{bbding}
\usepackage{lipsum}

\usepackage{xspace}

\newfloatcommand{capbtabbox}{table}[][\FBwidth]

\newcommand{\draftonly}[1]{#1}
\newcommand{\eat}[1]{}
\renewcommand{\draftonly}[1]{}
\definecolor{darkgreen}{RGB}{0, 102, 0}

\crefformat{section}{\S#2#1#3}

\makeatletter
\renewcommand{\@makefnmark}{}%
\renewcommand{\@makefntext}[1]{#1}%
\makeatother

\author[1]{Zijian Song}
\author[1]{Sihan Qin}
\author[2]{Tianshui Chen}
\author[1]{Liang Lin}
\author[1]{Guangrun Wang}

\author{\textbf{\normalsize{Zijian Song$^{1}$, Sihan Qin$^{1}$, Tianshui Chen$^{3,4}$, Liang Lin$^{1,2,3}$, Guangrun Wang$^{1,2,3*}$}} \\
$^1$Sun Yat-sen University\\
$^2$Guangdong Key Laboratory of Big Data Analysis and Processing\\
$^3$X-Era AI Lab\\
$^4$Guangdong University of Technology\\
Emails: \texttt{\{songzj8,qinsh9\}@mail2.sysu.edu.cn, wanggrun@gmail.com, linliang@ieee.org, chentianshui@gdut.edu.cn}\\
Website: \href{https://hcplab-sysu.github.io/PhysicalAutoregressiveModel/}{https://hcplab\-sysu.github.io/PhysicalAutoregressiveModel/}\footnote{$^*$corresponding author.}
}

\begin{abstract}
The scarcity of manipulation data has motivated the use of pretrained large models from other modalities in robotics. In this work, we build upon autoregressive video generation models to propose a Physical Autoregressive Model (PAR), where physical tokens combine frames and actions to represent the joint evolution of the robot and its environment.
PAR leverages the world knowledge embedded in video pretraining to understand physical dynamics without requiring action pretraining, enabling accurate video prediction and consistent action trajectories.
It also adopts a DiT-based de-tokenizer to model frames and actions as continuous tokens, mitigating quantization errors and facilitating mutual enhancement.
Furthermore, we incorporate a causal mask with inverse kinematics, parallel training, and the KV-cache mechanism to further improve performance and efficiency.
Experiments on the ManiSkill benchmark show that PAR achieves a 100\% success rate on the PushCube task, matches the performance of action-pretrained baselines on other tasks, and accurately predicts future videos with tightly aligned action trajectories.
These findings underscore a promising direction for robotic manipulation by transferring world knowledge from autoregressive video pretraining.
\end{abstract}

\begin{document}

\maketitle

\renewcommand{\floatpagefraction}{0.8}

\section{Introduction}
\label{sec:intro}








The success of generative pretraining in computer vision (CV) and natural language processing (NLP) has highlighted the power of large-scale pretraining, demonstrating strong generalization ability across diverse tasks~\cite{brown2020language, kirillov2023segment, tian2024visual, dubey2024llama, guo2025deepseek, SD-MVS}.
This paradigm inspires the robotics community to pursue powerful foundational models for visuomotor control and manipulation.
Yet, unlike CV and NLP, generative action pretraining presents unique challenges, as acquiring large-scale human demonstration data is both time-consuming and labor-intensive~\cite{zitkovich2023rt}. A promising strategy is to transfer knowledge from existing pretrained models to the action domain~\cite{zitkovich2023rt}. For instance, recent approaches build action policies upon large language models (LLMs), using action decoders to map linguistic outputs to low-level robot control, namely Vision-Language-Action-Models (VLAs)\cite{kim2024openvla, li2024cogact, bu2025agibot, zhao2025cot}.

However, the intrinsic gap between text modality and action modality poses significant challenges for such approaches, often resulting in suboptimal alignment between symbolic reasoning and physical control.
A more suitable solution is to build on pretrained video generation models, particularly autoregressive video generation models~\cite{gao2024vid, xie2025progressive}. These models predict future frames iteratively based on past observations, implicitly grounded in the physical dynamics of the physical world.
This predictive capability aligns naturally with the goal of action generation, as both tasks require understanding how the current state evolves over time.
Moreover, integrating actions with video representations creates a bidirectional synergy.
Actions ensure the predicted dynamics remain consistent with controllable outcomes.
Video offers valuable guidance for action generation through inverse kinematics.

\begin{figure}[t]
    \centering
    \includegraphics[width=0.6\linewidth]{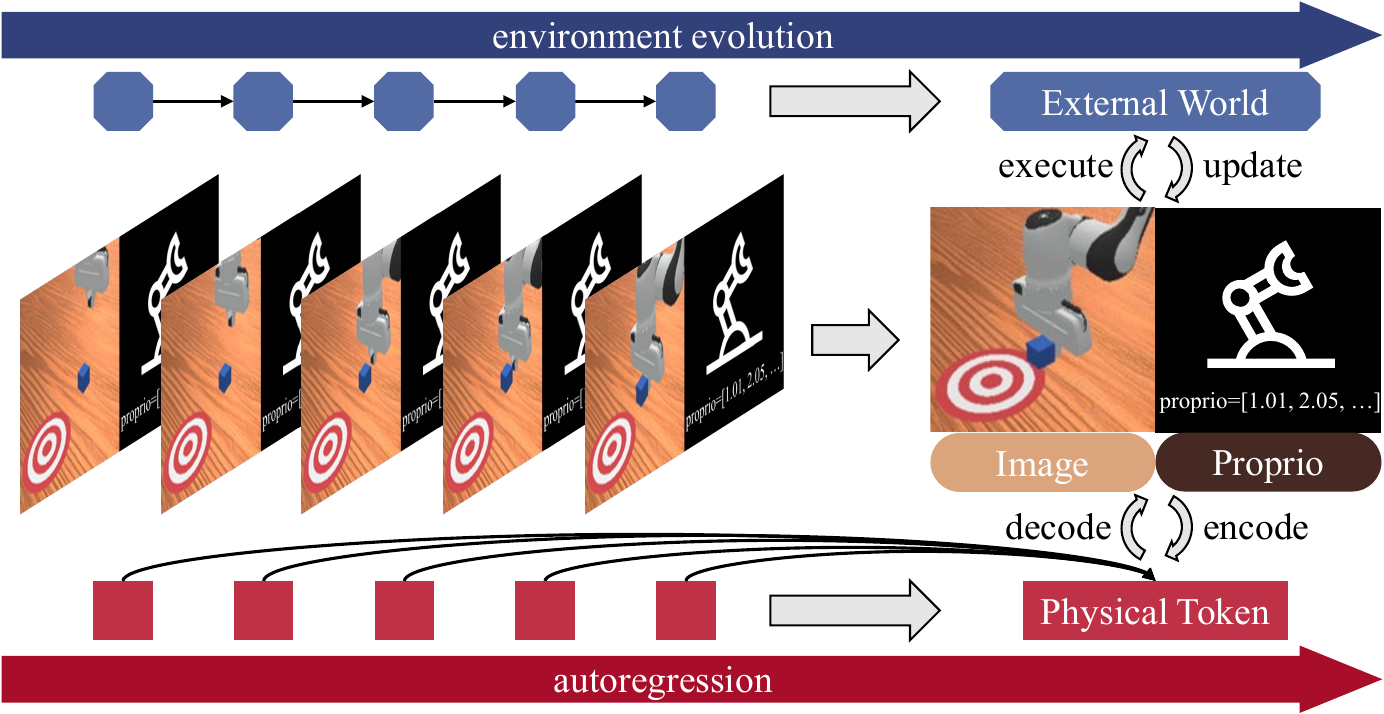}
    \caption{\textbf{The illustration of Physical Autoregression.} Our autoregressive process operates over a sequence of physical tokens (marked by red), each combining the visual world state (marked by orange) and the embodiment state (marked by black), providing a progressive estimate of their joint evolution.
    This process runs in sync with the environment: at each step, the predicted token is decoded into an image and an action, which interact with the environment to update its state (marked by blue), while the resulting observations and proprios are encoded back into the context.}
    \label{fig:world_evolve}
\end{figure}

The autoregressive nature provides an ideal foundation for representing the physical world.
Based on this insight, this work presents the \emph{Physical Autoregressive Model, PAR}. The underlying motivation is shown in~\cref{fig:world_evolve}.
Specifically, the frame tokens and the action tokens are combined into physical tokens, which effectively represent the joint evolution of robotic manipulation and environmental feedback.
To mitigate the scarcity of human demonstrations, we transfer world knowledge from video pretraining into our PAR model, facilitating a seamless transition from understanding visual dynamics to capturing physical dynamics.

A critical component for autoregressive models lies in the token representation~\cite{yu2023language, vuong2025action}.
Most existing methods rely on discrete token representations to represent visual signals and action signals~\cite{zitkovich2023rt, kim2024openvla, wang2025unified}.
This may introduce resolution errors whose accumulation propagates through long-horizon prediction and produces substantial trajectory drift\cite{zhang2025chain}.
Recent studies have explored continuous token representations for vector modeling~\cite{li2024autoregressive, deng2024autoregressive}, but their integration into a unified vision-action autoregressive framework remains limited.
In this work, we propose to represent both frame and action as continuous signals. Specifically, we leverage the Diffusion-Transformer model (DiT)~\cite{peebles2023scalable} coupled with Diffusion Loss training objective to model the arbitrary distribution of the continuous frame and action tokens. This design not only improves smoothness and coherence, but also facilitates a deeper mutual interaction between the continuous spaces of vision and action.

Besides, we further refine our Physical Autoregressive Model through two key components.
(1) During autoregression, we let the action tokens attend to the current frame tokens, which implicitly encode the prediction of the next visual state. This mechanism functions as an implicit inverse-kinematics step that improves accuracy.
To support this, we introduce a tailored causal masking strategy specifically designed for the frame–action joint autoregressive model.
(2) Similar to LLMs, we incorporate parallelized training~\cite{vaswani2017attention} to accelerate convergence and the KV-caching mechanism~\cite{NVIDIA2023LLM} for efficient inference.

We conduct experiments on the ManiSkill Benchmark~\cite{taomaniskill3}, which demonstrates the effectiveness of PAR.
Our method achieves a 100\% success rate on the PushCube task and performs on par with action-pretrained baselines on other tasks. Visualizations further show that PAR accurately predicts future video frames while generating action trajectories that remain tightly aligned with the predicted dynamics.
Ablation studies confirm the necessity of each key design component.

Key contributions of this paper includes:
(1) We combine frames and actions into physical tokens to construct a Physical Autoregressive Model. The world knowledge is inherited from video generation models to capture the iterative dynamics of the physical world, which needs zero action pretraining.
(2) We represent frames and actions as continuous vectors and model their distribution through DiT model and diffusion loss, allowing for fine-grained representation and deeper interaction between two modalities.
(3) We further incorporate a causal mask with implicit inverse-kinematics, parallel training, and KV-cache mechanism. Experiments on the ManiSkill benchmark validate the effectiveness of our overall framework as well as each of its key components.
\section{Related Work}
\label{sec:related_work}

\begin{figure}[t]
    \centering
    \begin{minipage}[b]{0.8\linewidth}
        \centering
        \includegraphics[width=1.0\linewidth]{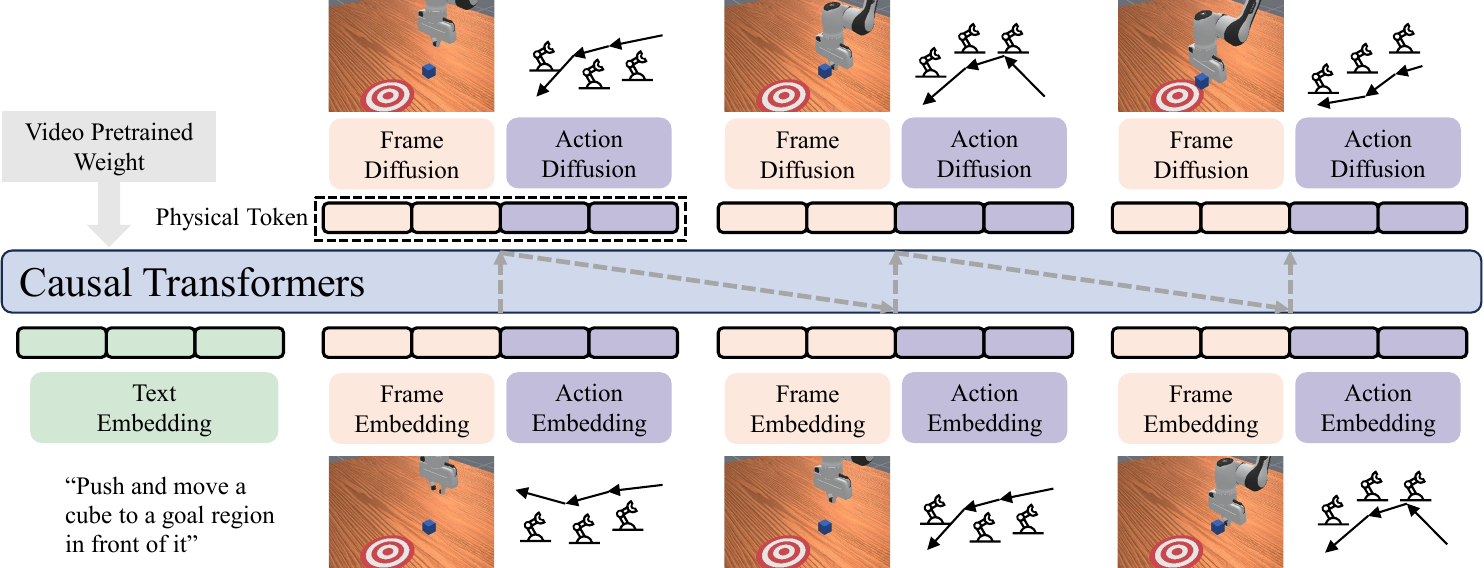}
    \end{minipage}
    \hfill
    \begin{minipage}[b]{0.18\linewidth}
        \centering
        \includegraphics[width=1.0\linewidth]{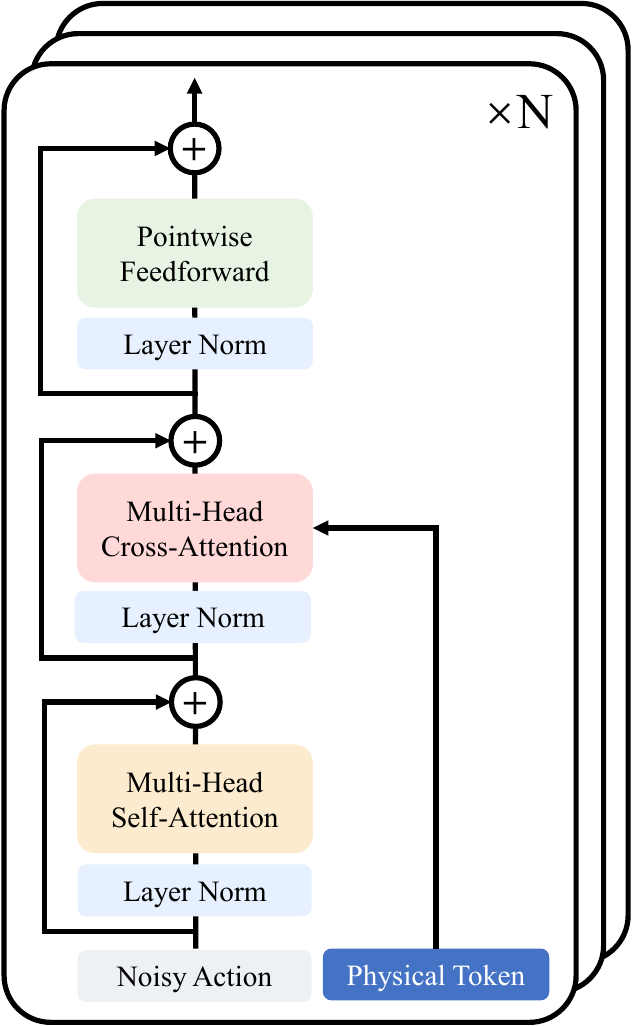}
    \end{minipage}
    \caption{\textbf{The model architecture of Physical Autoregressive Model.} Starting from text tokens, PAR leverages a causal Transformer to autoregressively predict physical tokens. Notably, we apply a frame diffusion and an action diffusion to estimate the conditional distributions of visual and action signals in continuous space.
    The structure of the action diffusion network is shown on the right column, where the predicted token serves as the conditional vector and is injected through cross-attention.
    }
    \label{fig:ar_transformer}
\end{figure}

\subsection{Vision-Language-Action Models}
The scarcity of manipulation demonstrations makes large-scale action pretraining prohibitively expensive. A recent trend in robotics involves transferring pretrained knowledge from LLMs to the action domain~\cite{zitkovich2023rt, kim2024openvla, black2024pi0, xu2025a0}. 
These approaches typically append a lightweight action head to the LLM, mapping language tokens to action outputs~\cite{wen2025tinyvla, wen2025dexvla, liu2025hybridvla, bu2025univla}.
They have demonstrated strong manipulation success rates with minimal or even zero action pretraining.
In contrast, our approach builds the action model on top of a pretrained video generation model rather than an LLM.
This design enables a seamless transfer of physical knowledge that aligns actions with the world dynamics captured by the video.

\subsection{Video-Action Joint Predition}

The concept of jointly predicting video and actions has been explored in prior studies, demonstrating its effectiveness~\cite{wu2023unleashing, cheang2024gr, cen2025worldvla, yelatent, fan2025towards}.
For example, UVA~\cite{li2025unified} jointly learns a video-action representation to model the correlation between video and action. VPP~\cite{hu2024video} leverages the pretrained video diffusion model for implicit inverse dynamics.
Building on this idea, our work employs an autoregressive framework that models the coupled evolution of the world and the robot in a step-by-step manner.
A key advantage is that our framework can leverage a variable-length history context for coherent planning and adaptive adjustment.

\subsection{Continuous Signal Tokenization}
For continuous signals such as images or actions, discretizing them may disrupt their inherent structure and introduce resolution errors~\cite{zhang2025chain}.
To preserve continuity for action generation, some approaches employ linear or MLP projectors but fall short for generative modeling~\cite{wu2023unleashing, fu2024context, zhang2025chain}.
To overcome this, Li et al.~\cite{li2024autoregressive} proposed using a denoising process to compute the conditional distribution of signals, agnostic to the signal distribution. This idea has been extended to video generation~\cite{deng2024autoregressive} and action prediction~\cite{pi05vision, li2025unified, team2024octo, zhong2025dexgraspvla}.
However, its application to action autoregressive modeling remains underexplored. In this work, we leverage a denoising process as a de-tokenizer for continuous action autoregression, fostering interactions between action and frame modalities.
\section{Preliminaries}
\label{sec:preliminaries}

This section provides a review of the key concepts that form the foundation of our work. We begin with an overview of diffusion models, then describe how the diffusion process can be adapted as a loss function for autoregressive modeling in a continuous-valued space. Finally, we discuss how these principles are applied to video generation.

\subsection{Diffusion Models}  
Diffusion Models\cite{ho2020denoising} are a class of generative models that produce samples by progressively adding Gaussian noise to clean data \(x_0\) and then learning to invert this noising process. Formally, the forward diffusion process is defined as:
\begin{equation}
q(x_t \mid x_{t-1})
= \mathcal{N}\bigl(x_t;\sqrt{1-\beta_t}\,x_{t-1},\,\beta_t I\bigr),
\end{equation}
where \(x_t\) is the noisy sample at timestep \(t\), \(\beta_t\) is the pre-defined noise schedule, and \(\epsilon\sim\mathcal{N}(0,I)\) is the injected noise. The reverse diffusion process is defined as:
\begin{equation}
p_\theta(x_{t-1} \mid x_t)
= \mathcal{N}\bigl(x_{t-1};\mu_\theta(x_t,t),\,\Sigma_\theta(x_t,t)\bigr),
\end{equation}
which inverts the corruption, starting from the standard normal prior \(x_T\sim\mathcal{N}(0,I)\) at the final timestep \(T\). Training then teaches the model to remove noise by minimizing
\begin{equation}
L_{\text{simple}}(\theta) = \mathbb{E}_{t,x_{0},\epsilon}\left[ \left\| \epsilon - \epsilon_{\theta}(\sqrt{\overline{\alpha}_{t}}x_{0} + \sqrt{1 - \overline{\alpha}_{t}}\epsilon, t) \right\|^{2} \right],
\end{equation}
where $\epsilon_{\theta}(\sqrt{\overline{\alpha}_{t}}x_{0} + \sqrt{1 - \overline{\alpha}_{t}}\epsilon, t)$ is the neural network's prediction of the added noise . At inference, the reverse process is applied to pure Gaussian noise to progressively remove noise and thus generate high-quality samples.

\subsection{Diffusion Loss}
Diffusion Loss~\cite{li2024autoregressive} repurposes the diffusion framework to model each continuous token’s conditional distribution \(p(x|z)\) by applying the noise–denoise cycle directly in latent space. Concretely, given a ground-truth token embedding \(x\in\mathbb{R}^{d}\) and its autoregressive context \(z\), we train a lightweight denoiser \(\epsilon_{\theta}(x_{t}|t,z)\) to predict the injected noise by minimizing the squared-error loss:
\begin{equation}
\mathcal{L}(z,x)=\mathbb{E}_{\epsilon,t}[||\epsilon-\epsilon_{\theta}(x_{t}|t,z)||^{2}].
\label{eq:diffusion_loss}
\end{equation}

Here, \(\epsilon\in\mathbb{R}^{d}\) is a noise vector sampled from \(\mathcal{N}(0,I)\). The noise-corrupted vector is \(x_{t} = \sqrt{\overline{\alpha}_{t}}x+\sqrt{1-\overline{\alpha}_{t}}\epsilon\), where \(\overline{\alpha}_{t}\) defines a noise schedule and \(t\) is a time step of that schedule. The notation \(\epsilon_{\theta}(x_{t}|t,z)\) indicates that the denoising network takes the corrupted token \(x_t\) as input and is conditioned on both the time step \(t\) and the context vector \(z\).

This loss enables us to learn \(p(x|z)\) directly in continuous space—eliminating the need for vector quantization. At inference, each token is sampled by running the learned reverse diffusion chain:
\begin{equation}
x_{t-1}=\frac{1}{\sqrt{\alpha_{t}}}\left(x_{t}-\frac{1-\alpha_{t}}{\sqrt{1-\overline{\alpha}_{t}}}\epsilon_{\theta}(x_{t}|t,z)\right)+\sigma_{t}\delta.
\end{equation}

Starting with a random vector \(x_{T}\sim\mathcal{N}(0,I)\), this iterative procedure produces a final sample \(x_{0}\sim p(x|z)\), yielding high-quality continuous embeddings.

This objective teaches \(\epsilon_\theta\) to invert the forward corruption, effectively learning \(p(x_n\mid z_n)\). At inference, one samples \(x_n\) by running the learned reverse diffusion chain from \(x_n^T\sim\mathcal{N}(0,I)\) back to \(x_n^0\), yielding high-quality continuous tokens without vector quantization.

\subsection{Autoregressive Modeling for Video}

The principle of autoregressive modeling in a continuous, non-quantized space can be extended to video generation~\cite{deng2024autoregressive}.

First, the temporal autoregressive process can be formally expressed by the conditional probability:
\begin{equation}
p(P,m,B,S_{1},...,S_{N})=\prod_{n=1}^{N}p(S_{n}|P,m,B,S_{1},...,S_{n-1}).
\end{equation}
In this formulation, the probability of generating the n-th frame, \(S_{n}\), is dependent on all prior frames (\(S_{1},...,S_{n-1}\)) and other conditioning contexts, ensuring coherence.

Second, the spatial autoregressive model generates each frame in a set-by-set order.
Concretely, the generation of token sets for the f-th frame is formulated as follows:
\begin{equation}
p(S_{n}^{\prime},S_{(n,1)},...,S_{(n,K)})=\prod_{k=1}^{K}p(S_{(n,k)}|S_{n}^{\prime},S_{(n,1)},...,S_{(n,k-1)}).
\end{equation}
Here, the k-th set of tokens, \(S_{(n,k)}\), is predicted conditioned on indicator features \(S_{n}^{\prime}\) and all previously generated sets (\(S_{(n,1)},...,S_{(n,k-1)}\)). This allows the model to leverage bidirectional context for generating spatial details.
\section{Method}
\label{sec:cmast_method}
\subsection{Problem Formulation}
As an autoregressive model, one key characteristic of our Physical Autoregressive Model (PAR) is its ability to operate with a variable-length history context.
This design enables the model to consistently follow its planned trajectory while dynamically adjusting its outputs as the context evolves.

Formally, given the task instruction $T$, we maintain an accumulated history context consisting of a sequence of image observations $ \left\{ O_0, O_1, \cdots, O_{N-1} \right\} $ and the corresponding sequence of action chunks $ \left\{ A_1, A_2, \cdots, A_{N-1} \right\} $.
The image sequence is one step longer than the action sequence because an initial observation exists before any action is taken.
The objective of PAR is to predict the subsequent image frame $ O_N $ and its associated action chunk $ A_N $ in an autoregressive manner.
The subscript $n$ denotes the timestep.

To enhance consistency, we adopt action-chunk modeling~\cite{chi2023diffusion, zhao2023learning}, where each action chunk $A_n$ consists of $L$ consecutive actions. 
Consequently, the action sampling rate is $L$ times that of the images.

\begin{figure}[t]
    \centering
    \includegraphics[width=0.4\linewidth]{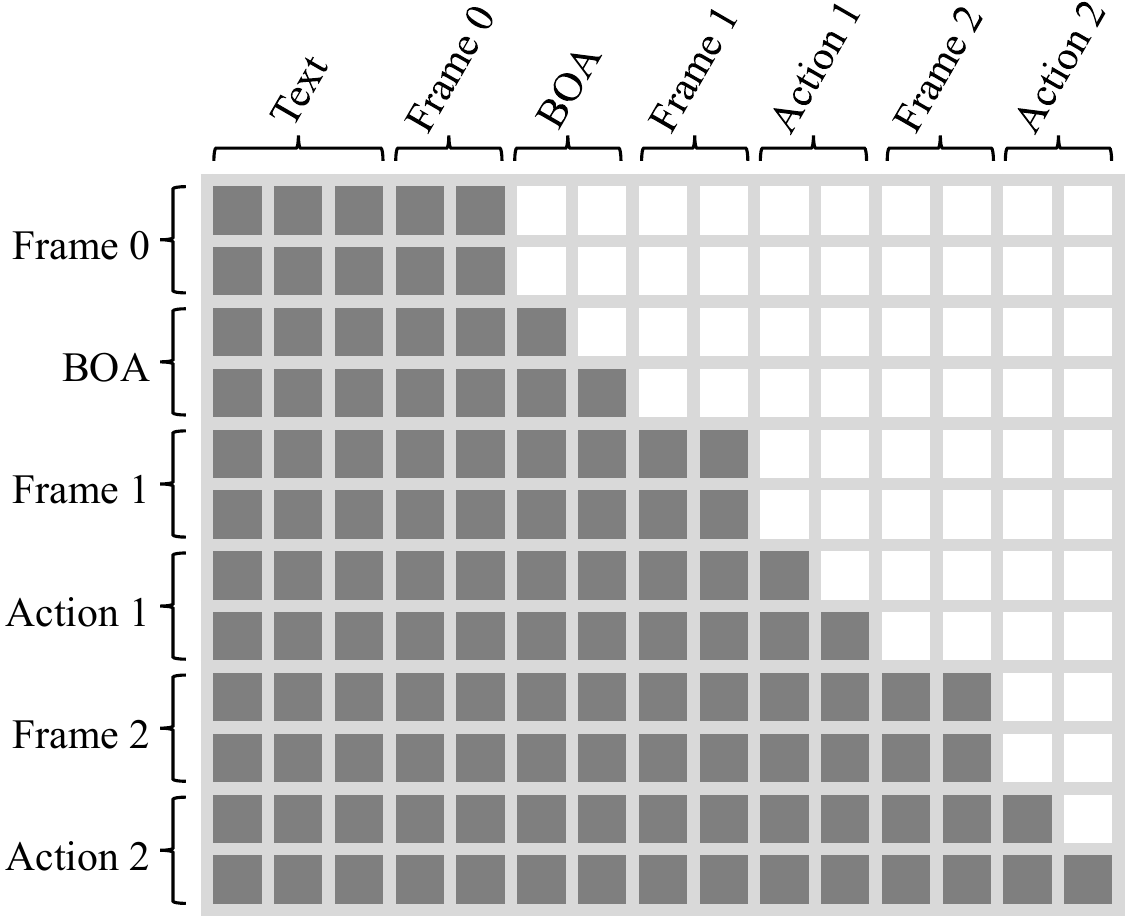}
    \caption{\textbf{The causal attention mask.} For illustration, we use 3 blocks to represent the task instruction tokens, 2 blocks to represent the frame tokens, 2 blocks to represent the action tokens.}
    \label{fig:casual_mask}
\end{figure}

\subsection{Model Architecture}

In this work, we construct our Physical Autoregressive Model (PAR) on top of a pretrained autoregressive video generation model, i.e., NOVA~\cite{deng2024autoregressive}. By transferring the world knowledge embedded in NOVA’s pretrained representations, PAR acquires a seamless understanding of physical dynamics without action pretraining, enabling it to unify video prediction and action generation within a single autoregressive framework.

\subsubsection{Tokenizer}
To fully exploit the capabilities of NOVA, our framework adopts its pretrained language and image encoders without modification.
Specifically, the task instruction is embedded into $K_t$ tokens, each with $d$-dimension features, using a frozen Phi model~\cite{javaheripi2023phi} followed by a linear projector.
For visual input, we employ an open-source 3D variational autoencoder~\cite{zheng2024open} to encode the video frames to the latent space, which is also kept frozen.
The resulting latent images are then projected to physical embedding space and flattened to form $K_O$ tokens, each with a feature dimension of $d$.
In addition, action chunks are encoded through a lightweight multi-layer perceptron (MLP), which maps continuous action signals into the same physical embedding space as the frame tokens.
Each action step constitutes a $d$-dimensional token, resulting in a total of $K_A = L$ tokens.

\begin{table}
    \centering
    \setlength{\tabcolsep}{8pt}
    \begin{tabular}{cc}
      \toprule
      Module & Parameters \\
      \midrule
Action-Tokenizer & 0.6M \\
Action-DeTokenizer & 21.1M \\
Frame-DeTokenizer & 32.8M \\
Casual-Transformer & 608.7M \\
      \midrule
Total & 663.2M \\
      \bottomrule
    \end{tabular}
    \caption{\textbf{PAR parameter count.} The PAR architecture only introduces minimal parameters to transfer knowledge from video pretraining to action generation.}
    \label{tab:param}
\end{table}

\subsubsection{Physical Autoregression}
We follow the standard \emph{token-by-token} autoregressive paradigm commonly used in large language models (LMMs).
The key difference is that our model operates on physical tokens rather than language tokens, which unifies frame and action signals.
This design allows us to jointly model the iterative evolution of both the robot and the external world.
Specifically, each physical token is constructed by concatenating a frame token with its corresponding action chunks in the token dimension, $P_{n} = \left[ O_n; A_n\right]$.
For the first observation frame, we concatenate it with a learnable Begin-Of-Action token, $A_0 = [BOA]$.
The autoregression process then predicts the next physical token conditioned on the sequence of preceding physical tokens.
This process can be written as:
\begin{equation}
    p\left( T,O_0,A_0,\cdots ,O_N,A_N \right)=p\left( T,P_0,\cdots ,P_N \right)=\prod_{n=0}^N{p\left( P_n|T,P_0,\cdots ,P_{n-1} \right)}.
\end{equation}
where the conditional distribution is modeled by a Transformer that exactly replicates the architecture of NOVA\cite{deng2024autoregressive}.

\subsubsection{De-Tokenizer}
Previous works typically rely on discrete tokenization for both images and actions~\cite{zitkovich2023rt, zheng2024open, wang2025unified}. This aligns well with language modeling, as discrete tokens can be directly integrated into the vocabulary.
However, such discretization introduces resolution errors that accumulate over the long horizon~\cite{zhang2025chain}.
Another line of work leverages linear or MLP projectors for decoding~\cite{wu2023unleashing, fu2024context, zhang2025chain}.
While these approaches avoid quantization artifacts, they primarily model deterministic values, limiting their generative capacity. Other extensions such as Gaussian Mixture Models still fail to capture arbitrary distributions~\cite{fu2024context, cheang2024gr}.

In this work, we address these limitations by adopting a continuous de-tokenizer, which maintains both a continuous and generative representation.
Moreover, our design also stimulates mutual enhancement between the image and action modalities.
Specifically, we follow the Diffusion-based de-tokenizer proposed by MAR\cite{li2024autoregressive}.
The tokens output by Transformer are taken as the conditional vector $Z_n$ for denoising process:
\begin{equation}
Z_n = Transformer(T, P_0, \cdots, P_{n-1}).
\end{equation}
A DiT network is applied to compute the probability of the predicted tokens $p(P_n|Z_n)$ through the Diffusion Loss:
\begin{equation}
\mathcal{L}(Z_n,P_n)=\mathbb{E}_{\epsilon,t}[||\epsilon-\epsilon_{\theta}(P_{n,t}|t,Z_n)||^{2}],
\end{equation}
where the $Z_n$ represents the conditioning vector.
The gradient is backpropagated to $z$ for optimizing the parameters in the Transformer backbone.

The decoding of frame tokens and action tokens is performed separately.
For frame tokens, we follow the same process used in NOVA, employing an additional autoregression process for masked patch set prediction.
For action tokens, we employ a lightweight DiT network for denoising, named Action-DiT module, where the conditioning vectors are injected into the DiT blocks via cross-attention.

\subsubsection{Causal Mask}
Our causal masking scheme differs from those commonly used in language models or video prediction models; the diagram can be found in~\cref{fig:casual_mask}. 
Each physical token consists of a frame part and an action part.
For the frame part, we employ a chunk-based attention mask, which allows all patches within the same frame to attend to each other.
For the action part, we use a temporal causal mask so that earlier actions within a chunk cannot attend to later actions.
Additionally, we allow action tokens to attend to frame tokens unidirectionally.
Since frame tokens processed by the Transformer are further decoded to future frames, this unidirectional design implicitly models inverse kinematics, enabling action planning to leverage future visual states for better accuracy.
Finally, across different time steps, we adopt a temporal causal mask to preserve temporal causality.

\subsubsection{Loss Function}
The training objective is defined as the average diffusion loss over the sequence of physical tokens, which is decomposed into frame and action components that are combined with equal weighting:
\begin{equation}
loss=\sum_{n=1}^N\mathcal{L}(Z_n,P_n)=\sum_{n=1}^N\left(\mathcal{L}_{obs}(Z_{O,n},O_n) + \mathcal{L}_{act}(Z_{A,n},A_n)\right).
\end{equation}

\subsubsection{Efficient Training and Inference}
For training, we adopt the standard parallelization strategy, where the entire token sequence is processed with teacher forcing and the loss for all tokens is computed in parallel. During inference, we employ a KV-cache mechanism~\cite{NVIDIA2023LLM} to store intermediate features at each layer, enabling efficient reuse  when generating subsequent tokens.


\begin{table}
    \centering
    \setlength{\tabcolsep}{2pt}
    \begin{tabular}{lcccc}
      \toprule
      Method & PushCube & PickCube & StackCube & Avg. \\
      \midrule
ACT [\citeyear{zhao2023learning}] & 76\% & 20\% & 30\% & 42\% \\
BC-T [\citeyear{mandlekar2021matters}] & 98\% & 4\% & 14\% & 39\% \\
DP [\citeyear{chi2023diffusion}] & 88\% & 40\% & \textbf{80}\% & 69\% \\
ICRT [\citeyear{vuong2025action}] & 77\% & \textbf{78}\% & 30\% & 62\% \\
RDT [\citeyear{liurdt}]  & \textbf{100}\% & 77\% & 74\% & \textbf{84}\% \\
      \midrule
PAR(Ours) & \textbf{100}\% & 73\% & 48\% & 74\% \\
      \bottomrule
    \end{tabular}
    \caption{\textbf{The success rate comparison} in the ManiSkill Benchmark. The best are highlighted by \textbf{bold}. Our method, without any action pretraining, achieves performance comparable to SOTA.}
    \label{tab:comparison_label}
\end{table}

\subsection{Implementation Details}
Our method requires no action pretraining, but it is finetuned on downstream manipulation tasks.
All finetuning experiments are conducted on a single NVIDIA A100-SXM-64GB GPU, with the longest training completed within 40 GPU hours.
During finetuning, we adopt full-parameter finetuning, updating all parameters of the NOVA along with the newly introduced action module.
The total number of learnable parameters is summarized in~\cref{tab:param}.
During training, we follow~\cite{li2024autoregressive} to sample $t$ by 4 times for each image and each action.
We employ the Rotary Positional Embeddings (RoPE)~\cite{su2024roformer} to preserve the temporal dependencies, with distinct frequency settings for frame and action tokens.

\section{Experiment}
\label{sec:experiment}

\begin{figure}[t]
    \centering
    \includegraphics[width=0.9\linewidth]{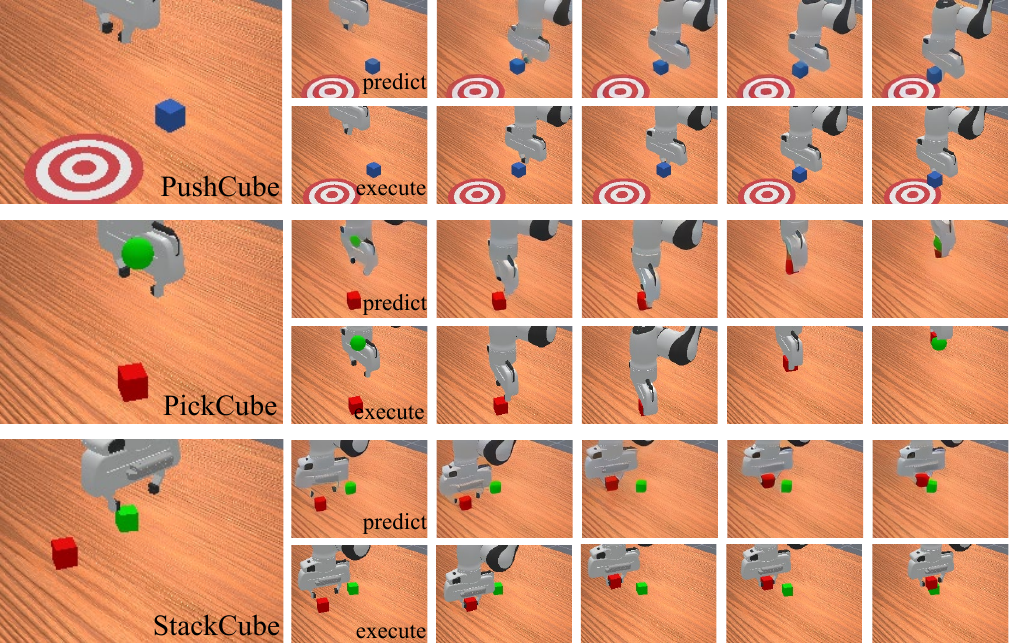}
    \caption{\textbf{Video predictions and actual task executions.} Each row shows PAR's predicted video alongside the corresponding execution video for three different tasks. The strong visual similarity between predicted and actual action videos highlights the effectiveness of our approach in transferring knowledge from video pretraining.}
    \label{fig:vis_success}
\end{figure}

\subsection{Evaluation Protocol}
\label{sec:evaluation_protocol}

\subsubsection{Benchmark}

We use the ManiSkill Benchmark~\cite{taomaniskill3} and follow prior works\cite{chi2023diffusion, li2025unified, cen2025worldvla} by evaluating the single task scenarios. ManiSkill Benchmark provides diverse, high-fidelity simulation environments and standardized metrics, making it a reliable platform for robotic manipulation evaluation.
We specifically focus on three manipulation tasks: PushCube, PickCube, and StackCube.
We render 1K demonstrates for each task as training data.

\subsubsection{Metric}
Each task is evaluated with 125 rollouts, covering 5 random seeds and 25 distinct initial states. We report the average success rate across these rollouts as the primary performance metric.

\subsubsection{Baseline}
We compared with the following baseline, all methods are fine-tuned on the same data as our model and tested using the same random seed and initial states.

\begin{itemize}
    \item \textbf{Action Chunk Transformer (ACT)}~\cite{zhao2023learning} leverages action chunking and encoder-decoder transformer layers to learn bimanual manipulation policies.
    \item  \textbf{Behavior Cloning Transformer (BC-T)}~\cite{mandlekar2021matters} directly learn the mapping from visual observations to actions through transformer architecture.
    \item \textbf{Diffusion Policy (DP)}~\cite{chi2023diffusion} reformulates action generation as a conditional denoising diffusion process and achieves SOTA visuomotor performance. We use the Transformer-based design from their original implementation for all tasks.
    \item \textbf{In-Context Robot Transformer (ICRT)}~\cite{fu2024context, vuong2025action} is pretrained on in-context action trajectories and uses a Lipschitz-conditioned VAE for action embedding.
    \item \textbf{RobotDiffusionTransformer (RDT)}~\cite{liurdt} is an end-to-end robotic model based on the DiT, which has 1.3 billion parameters and is pretrained on a large-scale robotic dataset.
\end{itemize}

\subsection{Quantitative Comparison}

We quantitatively evaluate our method across three manipulation tasks, with results summarized in~\cref{tab:comparison_label}.
On average, our method achieves the second-best success rate, only behind RDT, which requires action pretraining.

In the PushCube task, our approach achieves a 100\% success rate, significantly outperforming the ICRT~\cite{fu2024context} baseline by 30\%.
In the PickCube task, our performance is within 7\% of the SOTA baseline, RDT~\cite{liurdt}, despite RDT relying on action pretraining while our method does not.
In the StackCube task, our method falls behind DP~\cite{chi2023diffusion} and RDT~\cite{liurdt}, but still outperforms ICRT~\cite{fu2024context} by 60\%.
These results highlight the strength of our approach in transferring world knowledge from video pretraining, with zero action pretraining and just 30M additional parameters.

Overall, these results validate the advantages of our autoregressive joint modeling framework, which facilitates cross-modal interaction and offers a promising direction for addressing data scarcity in action pretraining.

\subsection{Ablation Study}
We conduct a ablation study to examine the contribution of each component in our framework. Specifically, we compare our full method, denoted as PAR-Full against two ablated variants: (1) PAR-NoAR, which removes the autoregressive architecture. (1) PAR-Discrete, which replaces the generative de-tokenizer with a discriminative one.
These ablation models are trained and evaluated using the same evaluation protocol as our full method (see ~\cref{sec:evaluation_protocol}).
The results are summarized in~\cref{tab:ablation}.
PAR-Full consistently outperforms both baselines by a large margin, confirming the effectiveness of our designs.

\subsubsection{Autoregressive Architecture}
We remove the entire autoregressive framework inherited from NOVA to investigate its contribution, remaining only the MLP-based action encoder and the diffusion-based action decoder.
Consequently, the model directly decodes the embedded visual tokens and text tokens without any autoregressive transformer reasoning.
This modification leads to a decrease in average success rate from 73.6\% to 11.2\%, demonstrating the critical role of the autoregressive module.
This result suggests that our autoregressive video pretraining enables joint reasoning over the robot and the environment, effectively providing implicit inverse kinematics guidance to the action decoder, and thereby improving execution accuracy.

\begin{table}
    \centering
    \setlength{\tabcolsep}{2pt}
    \begin{tabular}{lcccc}
      \toprule
      Method & PushCube & PickCube & StackCube & Avg. \\
      \midrule
PAR-NoAR  & 29.6\% & 4.0\% & 0.0\% & 11.2\% \\
PAR-Discrete & 87.2\% & 65.6\% & 7.2\% & 53.3\% \\
PAR-Full & \textbf{100.0}\% & \textbf{72.8}\% & \textbf{48.0}\% & \textbf{73.6}\% \\
      \bottomrule
    \end{tabular}
    \caption{\textbf{The ablation study} in the ManiSkill Benchmark. The best are highlighted by \textbf{bold}.}
    \label{tab:ablation}
\end{table}

\begin{figure}[t]
    \centering
    \includegraphics[width=0.6\linewidth]{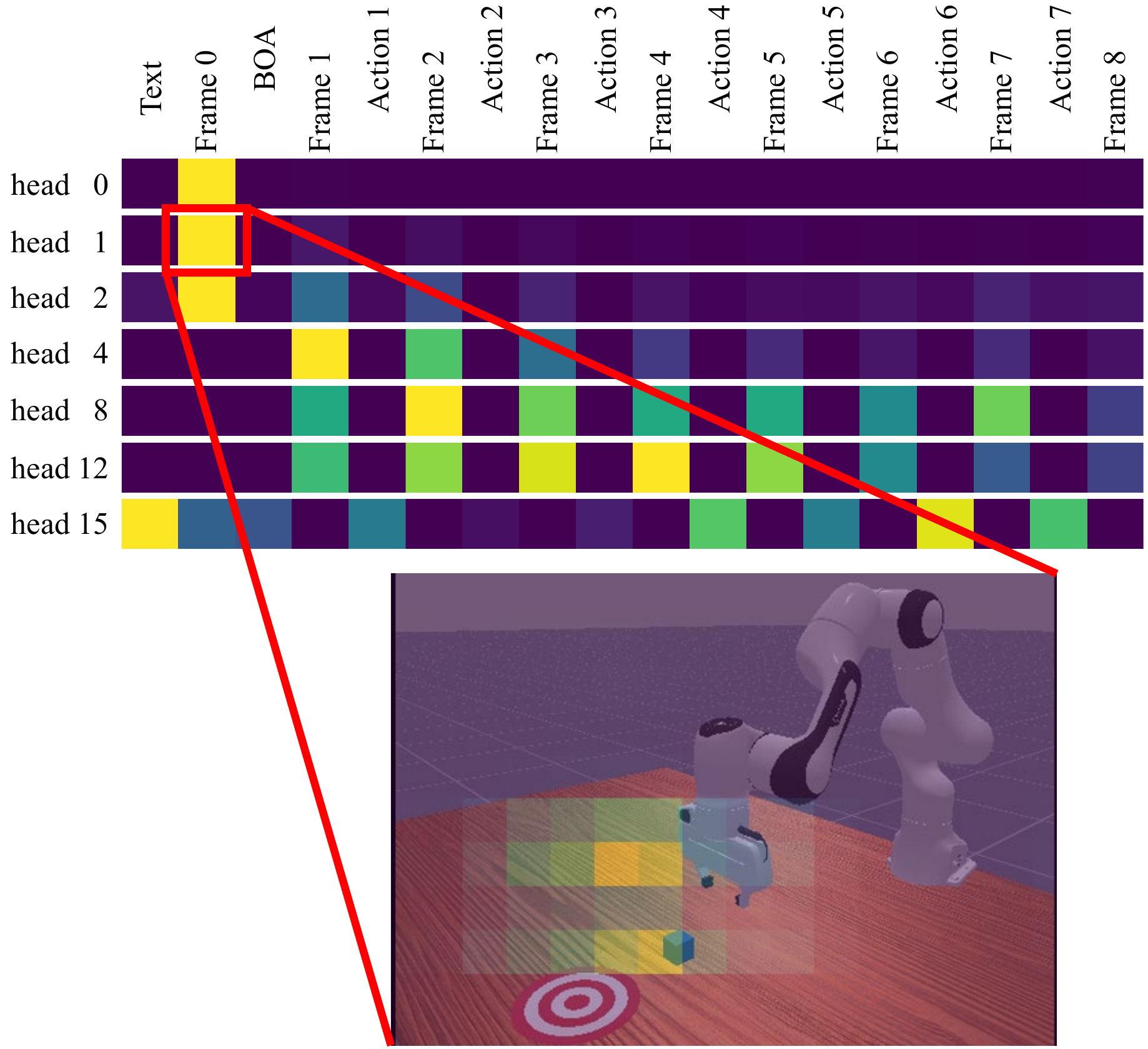}
    \caption{\textbf{The attention map.} The top row shows the token-level attention map, indicating how the predicted action attends to previous frame and action tokens. The bottom row shows the pixel-level attention map, indicating how the predicted action attends to different spatial regions of the frame.}
    \label{fig:attn_map}
\end{figure}

\subsubsection{De-Tokenizer Type}
To assess the role of the generative de-tokenizer, we replace the diffusion-based de-tokenizer with a MLP-based discriminative de-tokenizer, aligning with designs adopted in prior works~\cite{wu2023unleashing, fu2024context, zhang2025chain}.
This modification reduces the average success rate from 73.6\% to 53.3\%, indicating that the discriminative de-tokenizer may collapse to deterministic value regression and fail to capture uncertainty in the action distribution. This loss of distributional modeling likely results in overfitting and poorer generalization.

\subsection{Visualization}

\subsubsection{Video Action Alignment}
We visualize the predicted video sequences alongside the corresponding execution videos in~\cref{fig:vis_success}. The results reveal a clear alignment between the predicted visual trajectories and actual robot actions, with highly similar motion trajectories and timing of key actions.
For example, in the second row of the PickCube task, the predicted video correctly captures \textcolor{brown}{the rotation of the arm to match the cube’s orientation}, a meticulous step for a successful grasping.
The actual execution precisely mirrors this behavior, \textcolor{brown}{rotating by the correct angle to match the cube’s orientation} and enable a successful grasping.
This consistency highlights not only the model’s fine-grained scene understanding from video pretraining, but also its effectiveness in transferring that understanding into action planning.

\subsubsection{Attention Map}
We visualize the attention maps of PAR to better understand how it attends to visual and action cues during prediction, as shown in~\cref{fig:attn_map}. At the token level, different attention heads selectively focus on frame tokens or action tokens, indicating that the model effectively captures the interplay between perception and action for joint modeling. At the pixel level, the attention is concentrated on task-critical regions, such as the cube, target area, and the robotic arm, demonstrating that PAR can attend to spatially relevant features that directly influence task execution.

\subsubsection{Failure Case Study}
\begin{figure}[t]
    \centering
    \includegraphics[width=0.7\linewidth]{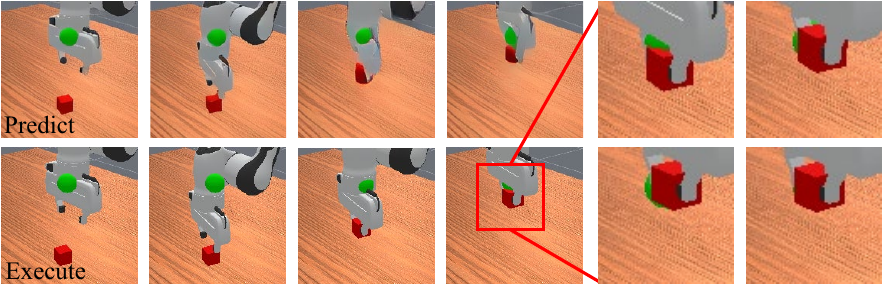}
    \caption{\textbf{A failure case} in the PickCube task. The top row shows the predicted video while the bottom row presents the actual execution video. The last two columns provide zoomed-in views of the final few steps of the execution. The robot arm fails to find the correct placement (marked by the green point) along the depth axis.}
    \label{fig:failure_case}
\end{figure}

We illustrate a failure case in~\cref{fig:failure_case}.
While the predicted trajectory seems valid, the robot misses the green target when moving the red cube.
Upon inspecting the last few steps of the manipulation, we observe that the robot arm performs light movements around the target area, failing to find the correct placement along the depth axis.
We hypothesis that this is because the ManiSkill Benchmark setup only provides single-view RGB observations, making it difficult to infer the depth information between the end-effector and the goal area. 
Future works could incorporate depth-estimation module (e.g. \cite{ze20243d, Yuan2025b, li2025bridgevla}) to enhance spatial awareness.

\section{Conclusion}
\label{sec:conclusion}

This work introduces the Physical Autoregressive Model (PAR), a unified framework that autoregressively models the coupled dynamics of visual observations and robotic actions. By transferring world knowledge from video pretraining, PAR captures physical dynamics without action pretraining, enabling accurate video prediction and coherent action generation. Our design leverages continuous token representations for both frames and actions, facilitating fine-grained modeling and cross-modal interaction.
Additionally, we incorporate a causal mask with implicit inverse kinematics, along with efficient training and inference strategies.
Our experiments on the ManiSkill benchmark show that PAR achieves a 100\% success rate on the PushCube task and matches the performance of action-pretrained baselines on other tasks.
It further demonstrates strong alignment between predicted video and action trajectories, evidencing successful transfer of world knowledge from video pretraining to action generation.
Ablation studies further validate the necessity of each component in our design.
Overall, by leveraging physical autoregression from video pretraining, our method suggests a promising new direction for addressing data scarcity in action pretraining.

\noindent{\textbf{Limitations.}}
In this work, we adopt full-parameter finetuning to showcase our approach. However, as a transformer-based model, PAR should support parameter-efficient finetuning like LoRA to achieve comparable performance at lower training costs, which we leave for future exploration.

\bibliography{main}

\end{document}